\documentclass[conference]{IEEEtran}
\IEEEoverridecommandlockouts
\usepackage{algorithm}%算法

\usepackage{multirow}

\usepackage{booktabs} %短横线

\usepackage{cite}
\usepackage{amsmath,amssymb,amsfonts}
\usepackage{algorithmic}
\usepackage{graphicx}
\usepackage{textcomp}
\usepackage{xcolor}
\def\BibTeX{{\rm B\kern-.05em{\sc i\kern-.025em b}\kern-.08em
    T\kern-.1667em\lower.7ex\hbox{E}\kern-.125emX}}
\begin{document}

\title{Two-stage optimized unified adversarial patch for attacking visible-infrared cross-modal detectors in the physical world

% {\footnotesize \textsuperscript{*}Note: Sub-titles are not captured in Xplore and
% should not be used}
% \thanks{Identify applicable funding agency here. If none, delete this.}
}

\author{\IEEEauthorblockN{1\textsuperscript{st} Chengyin Hu  \quad 2\textsuperscript{st} Weiwen Shi}
\IEEEauthorblockA{\textit{School of Computer Science and Engineering} \\
\textit{University of Electronic Science and Technology of China}\\
chengdu, China \\
cyhuuestc@gmail.com \quad weiwen\_shi@foxmail.com}}

\maketitle

\begin{abstract}
Currently, many studies have addressed security concerns related to visible and infrared detectors independently. In practical scenarios, utilizing cross-modal detectors for tasks proves more reliable than relying on single-modal detectors. Despite this, there is a lack of comprehensive security evaluations for cross-modal detectors. While existing research has explored the feasibility of attacks against cross-modal detectors, the implementation of a robust attack remains unaddressed. This work introduces the Two-stage Optimized Unified Adversarial Patch (TOUAP) designed for performing attacks against visible-infrared cross-modal detectors in real-world, black-box settings. The TOUAP employs a two-stage optimization process: firstly, PSO optimizes an irregular polygonal infrared patch to attack the infrared detector; secondly, the color QR code is optimized, and the shape information of the infrared patch from the first stage is used as a mask. The resulting irregular polygon visible modal patch executes an attack on the visible detector. Through extensive experiments conducted in both digital and physical environments, we validate the effectiveness and robustness of the proposed method. As the TOUAP surpasses baseline performance, we advocate for its widespread attention.
\end{abstract}


\section{Introduction}

With the advent of deep neural networks (DNNs), this technology has found application across diverse domains, achieving notable success in areas such as image classification \cite{ref1}, object detection \cite{ref2}, and semantic segmentation \cite{ref3}. In the realm of object detection, the majority of existing efforts \cite{ref4, ref38} concentrate on enhancing object detectors trained on samples captured in visible light environments (visible detectors), leading to significant advancements. However, a notable drawback is the sharp performance degradation or complete failure of detectors in dark environments. To address the challenge of target detection in low-light conditions, some studies \cite{ref6, ref7} utilize infrared sensors to capture infrared samples for training detectors (infrared detectors). Yet, the performance of infrared detectors can be susceptible to interference when the target temperature closely matches the ambient temperature. Consequently, in real-world scenarios, employing cross-modal detectors (visible-infrared detectors) \cite{ref8} for tasks such as autonomous driving and pedestrian detection can yield more robust performance.

\begin{figure}
\centering
\includegraphics[width=1\columnwidth]{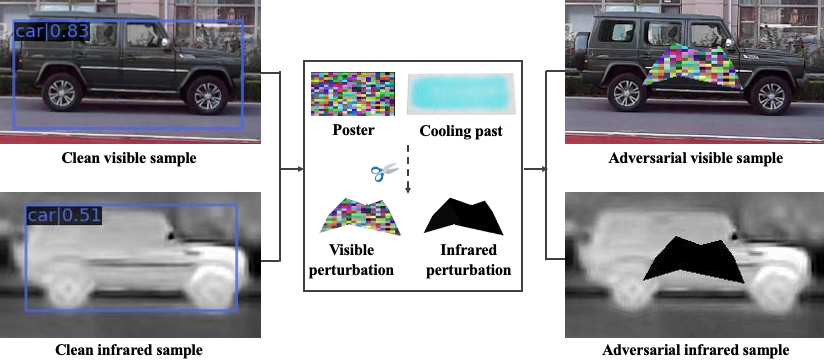} 
\caption{Unified adversarial patch. This illustration depicts the concise process of launching an attack on a vehicle detector using irregular polygonal patch.}
\vspace{-0.3cm}
\label{figure1}
\end{figure}

While DNNs-based tasks have made significant progress, adversarial attacks \cite{ref9} have emerged as a prominent focus for researchers. Adversarial attacks primarily fall into two main categories: digital attacks and physical attacks. In contrast to digital attacks \cite{ref10}, which target advanced DNNs by introducing slight perturbations to digital images, physical attacks \cite{ref17} require adding substantial perturbations to the target object. Subsequently, these perturbed objects are captured by a camera, and the resulting samples are input into DNNs to execute the attack. In the context of physical attacks against object detectors, the majority of studies \cite{ref12, ref13} have concentrated on exploring attacks against visible detectors. Recently, attention has gradually shifted towards attacks against infrared detectors in some works \cite{ref14, ref15}. However, physical attacks targeting a single modality are ineffective against cross-modal detectors. For instance, a physical attack against a visible detector may generate disturbances that are challenging for an infrared sensor to perceive, rendering it unsuccessful in attacking the infrared detector. Similarly, perturbations generated by physical attacks against infrared detectors are challenging to deceive visible detectors. Some works \cite{ref16, ref36} have proposed physical attacks against cross-modal detectors. While the patches generated by their method can successfully attack cross-modal detectors, achieving robustness remains a challenge.

Building upon the aforementioned discussions, this work introduces a Two-stage Optimized Unified Adversarial Patch (TOUAP) designed for black-box scenarios, as illustrated in Figure \ref{figure1}. The methodology involves a two-stage process to optimize physical parameters and create an adversarial patch. In the first stage, the infrared patch undergoes simulation and modeling to formalize its physical parameters. These parameters are optimized using the Particle Swarm Optimization algorithm \cite{ref37} to derive the optimal physical characteristics for the adversarial patch. Subsequently, in the second stage, simulation modeling is employed to create the visible patch. The color QR code serves as the physical parameters for the patch, optimized again using the Particle Swarm Optimization algorithm. The shape obtained in the first optimization stage is used as a mask, and this mask, combined with the color QR code patch, produces an irregular polygon QR code pattern—the visible patch. During the deployment phase, irregular polygonal color QR codes are printed as posters to create the physical visible patch. For the infrared patch, a cold patch, serving as cooling material (where the low-temperature material appears as a black pattern in the infrared sensor), is cut and affixed to the back of the poster. The resulting poster, featuring the irregular polygonal color QR code and pasted with the cold patch, constitutes the cross-modal unified adversarial patch. This patch is then applied to the target object, and photos or videos are captured using visible and infrared sensors to obtain the final physical sample.

The challenges inherent in executing physical attacks against cross-modal detectors primarily revolve around two key aspects: (1) Formulating irregular polygons and generating a unified adversarial patch; (2) Achieving a robust adversarial attack in a cross-modal context. To address challenge (1), our approach employs a grid partitioning technique to confine the positions of the irregular polygon's vertices. In practical physical experiments, we utilize an octagon for the attack, with each vertex constrained within the outer sub-square of a nine-palace grid to maximize its adversarial impact. The resulting shape of the infrared patch is amalgamated with the color QR code to create the unified adversarial patch. Regarding challenge (2), we employ the nine-palace grid to limit the irregular polygon's vertices and maximize its adversarial effect. Simultaneously, the color QR code serves as the visible light patch to further enhance the adversarial impact. Additionally, robust improvements are applied to the patches using the Expectation Over Transformation framework \cite{ref18}.

Our method demonstrates effectiveness and robustness throughout the process. In the first and second stages, we obtain the infrared and visible patches, respectively. The cold patch is then affixed to the back of the printed poster to produce the final unified adversarial patch. We validate the efficacy and robustness of our proposed method across two datasets (FLIR V2\_1 \cite{ref19} and LLVIP \cite{ref20}) in both single modality and cross-modality settings, specifically in physical environments. Unified adversarial patches are employed to evaluate the effectiveness and robustness of physical attacks. Notably, our method is straightforward to deploy, requiring only posters and cold patches as physical materials. The associated material costs are minimal, not exceeding \$5. The contributions of our work are summarized as follows:

\begin{itemize}

\item We introduce a Two-stage Optimized Unified Adversarial Patch (TOUAP) designed for black-box physical attacks in the visible-infrared cross-modal scenario. TOUAP employs particle swarm optimization to separately optimize infrared and visible patches, resulting in a robust unified adversarial patch for cross-modal attacks. Our proposed TOUAP is characterized by its effectiveness, robustness, cost-effectiveness, and easy deployment, presenting a significant security threat in the field of computer vision.

\item We employ a nine-square grid area to constrain the irregular octagon's fixed-point position, maximizing the adversarial impact. Utilizing a two-stage optimization strategy, we sequentially optimize the infrared and visible patches, obtaining the most adversarial patches in their respective modals. This approach ensures effective and robust cross-modal adversarial attacks.

\item Extensive experiments validate the effectiveness and robustness of TOUAP. In digital attacks, we conduct single-modal and cross-modal attacks on FLIR V2\_1 and LLVIP datasets, respectively, showcasing its effectiveness in digital environments. Deploying a unified adversarial patch in the physical realm effectively deceives visible and infrared detectors, demonstrating the efficacy of TOUAP in physical settings. Robustness analysis reveals TOUAP's superior attack success rate over the baseline. Additionally, we deploy TOUAP against various advanced detectors, achieving superior adversarial effects and confirming the robustness of the proposed approach.
 
\end{itemize}

\begin{figure*}
\centering
\includegraphics[width=0.7\linewidth]{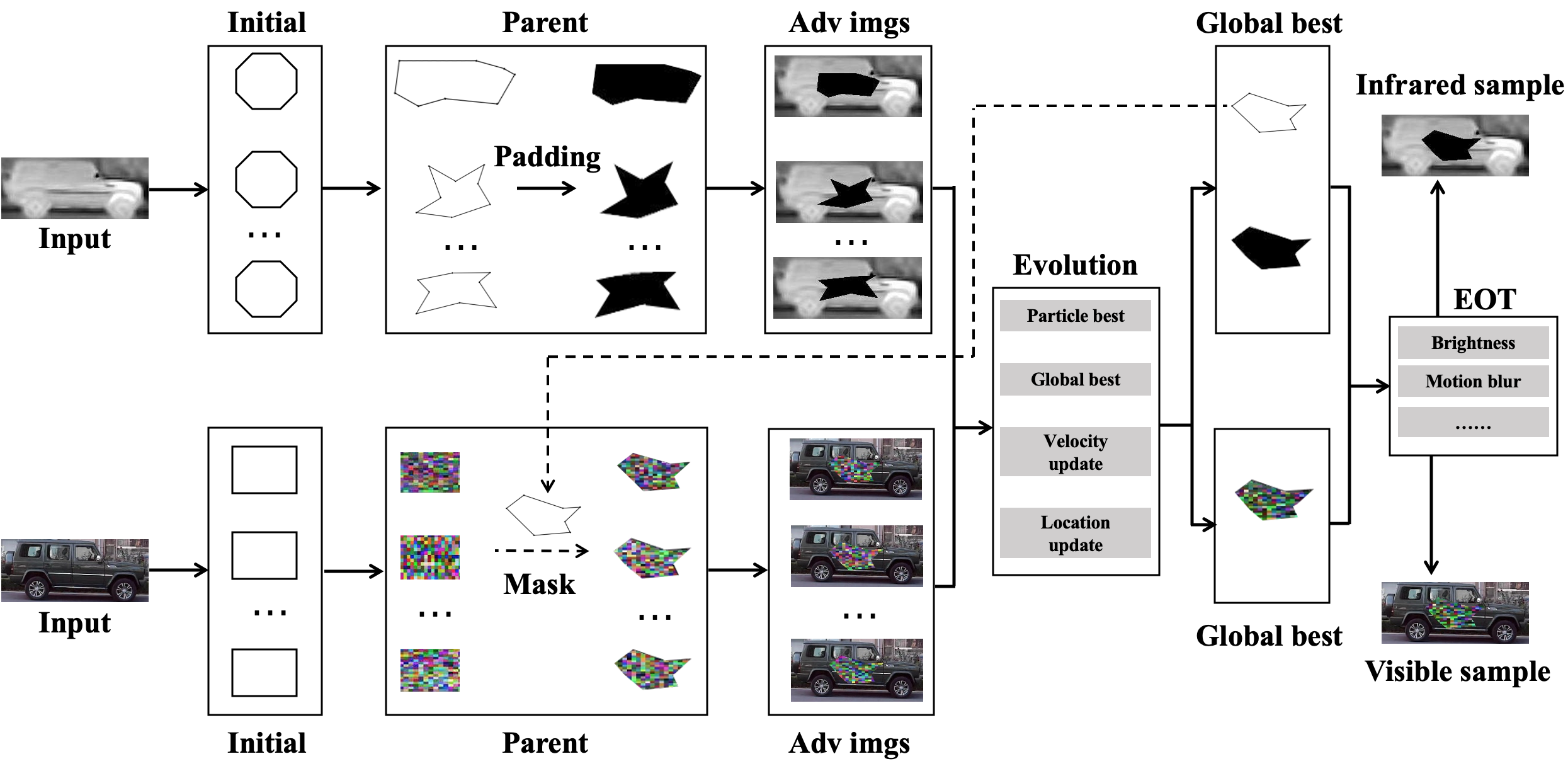} 
\caption{Summary graph of two-stage optimized unified adversarial patch. }
\vspace{-0.3cm}
\label{figure2}
\end{figure*}

\section{Related works}

The concept of adversarial attacks was initially introduced by Szegedy et al. \cite{ref9}, marking the beginning of extensive research in this domain \cite{ref21, ref22}. In this section, our emphasis is on adversarial attacks targeting object detectors.

\subsection{Adversarial attacks in the visible field}

Adversarial attacks against visible detectors have typically involved deploying adversarial patches onto the target object. In this context, certain approaches \cite{ref11, ref23} have investigated attacks on road sign detectors by affixing cartoon stickers to road signs or placing translucent stickers on camera lenses. This makes the physical perturbations generated less conspicuous to human observers but may compromise some level of robustness. Another set of approaches \cite{ref24, ref25, ref26, ref27, ref28} has focused on attacking pedestrian detectors by optimizing and printing adversarial patches, which are then applied to pedestrian clothing. Since pedestrian clothing constitutes non-rigid targets, these works bridge the gap between digital and physical attacks using techniques like EOT \cite{ref18} and TPS \cite{ref29} to enhance adversarial patches robustly, aiming for resilient physical attacks. Additionally, some works \cite{ref12, ref13, ref30} have delved into physical attacks against vehicle detectors, employing 3D rendering to generate textures or camouflage for adding perturbations to vehicles comprehensively. Notably, these methods concentrate on concealing the vehicle from visible detectors, irrespective of the perturbation's size. As vehicles are rigid targets, commonly employed robust optimization techniques include EOT \cite{ref18}, TV \cite{ref31}, etc.

\subsection{Adversarial attacks in the infrared field}

Zhu et al. \cite{ref14} introduced the Bulb attack, employing a bulb plate as a perturbation to launch an attack on infrared pedestrian detectors. The method utilizes a small bulb as a heating element, exhibiting white disturbance in the infrared sensor. Experimental verification confirms the effectiveness of this attack. Subsequently, they presented the QR attack \cite{ref15}, employing aerogel to create infrared "QR code coats" for multi-view attacks on infrared pedestrian detectors. Aerogel serves as a thermal insulation material, displaying black perturbation in the infrared sensor. The repeated superposition principle of the two-dimensional code pattern facilitates 2D optimization for multi-view attacks. To execute stealthy infrared attacks, Wei et al. \cite{ref34} introduced HCB, utilizing hot and cold patches as heating and cooling materials. These present white and black perturbation blocks in the infrared sensor. PSO \cite{ref37} is employed for optimizing the execution of black-box attacks. As it can be affixed inside clothing, human observers cannot detect the perturbation without the aid of an infrared sensor. More recently, Wei et al. \cite{ref35} proposed AIP, utilizing aggregation optimization to generate adversarial patches. Aerogel serves as a physical perturbation, which is cut and pasted onto pedestrians, enabling easily deployable physical attacks against infrared pedestrian detectors.

\subsection{Adversarial attacks in the visible-infrared field}

Kim et al. \cite{ref36} investigated physical attacks against visible-infrared pedestrian detectors. They employed sandpaper, steel, and aluminum as physical perturbations in the infrared environment, leveraging the distinct thermal emissivity characteristics of different metals for the attacks. In visible light environments, RGB perturbations were optimized using non-printable score loss and total variance loss. The perturbations in the two modals were then superimposed using overlay coverage to execute the physical attack. However, the method of covering the perturbed poster on the metal material plate made the infrared perturbation less conspicuous, weakening the robustness of the infrared attack. Additionally, the different shapes of the perturbation in the two modals were identified as drawbacks. Wei et al. \cite{ref16} explored Unified Adversarial Perturbation (UnfAP) for visible-infrared detectors. They employed an evolutionary algorithm to optimize the curvelet adversarial perturbation shape in a single stage, utilizing aerogel as a unified adversarial patch for an effective cross-modal physical attack. The perturbation was deemed insufficiently strong to execute a robust attack in the visible light environment, especially when utilizing a pure white color.In contrast, our method adopts a two-stage optimization strategy, allowing the perturbation in the two modals to maximize its adversarial effect. Furthermore, the use of a color QR code as the perturbation in the visible light environment enhances the robustness of our proposed method in the physical environment.

\section{Methodology}

\subsection{Problem definition}

In the cross-modal object detection tasks, we represent visible clean samples and adversarial samples as ${x}_{vis}$ and ${x}_{vis}^{adv}$, respectively. Similarly, infrared clean samples and adversarial samples are denoted as ${x}_{inf}$ and ${x}_{inf}^{adv}$. The detectors for visible and infrared are represented by ${f}_{vis}$ and ${f}_{inf}$ respectively. The primary objective of this work is to develop a unified adversarial patch capable of attacking both visible and infrared detectors, thereby achieving the following goals:

\begin{equation}
    \label{Formula1}
    max\{{f}_{vis}({x}_{vis}^{adv}), {f}_{inf}({x}_{inf}^{adv})\} < \delta
\end{equation}
where ${f}_{vis}({x}_{vis}^{adv})$ and ${f}_{inf}({x}_{inf}^{adv})$ denote the confidence scores for detecting the target in the visible and infrared detectors, respectively. The symbol $\delta$ represents the predefined confidence threshold.

A visual representation of our methodology is illustrated in Figure \ref{figure2}. The two-stage optimization procedure employs PSO \cite{ref37} to iteratively optimize the infrared and visible patches. To achieve a unified adversarial patch, the shape details of the most adversarial infrared patch obtained during the first stage are employed as a mask for the subsequent optimization of the visible patch. This process generates a visible patch with a shape matching that of the infrared patch. Subsequently, the patch undergoes robust enhancement using the EOT framework. Ultimately, a robust cross-modal adversarial patch is derived.

\begin{figure}
\centering
\includegraphics[width=1\linewidth]{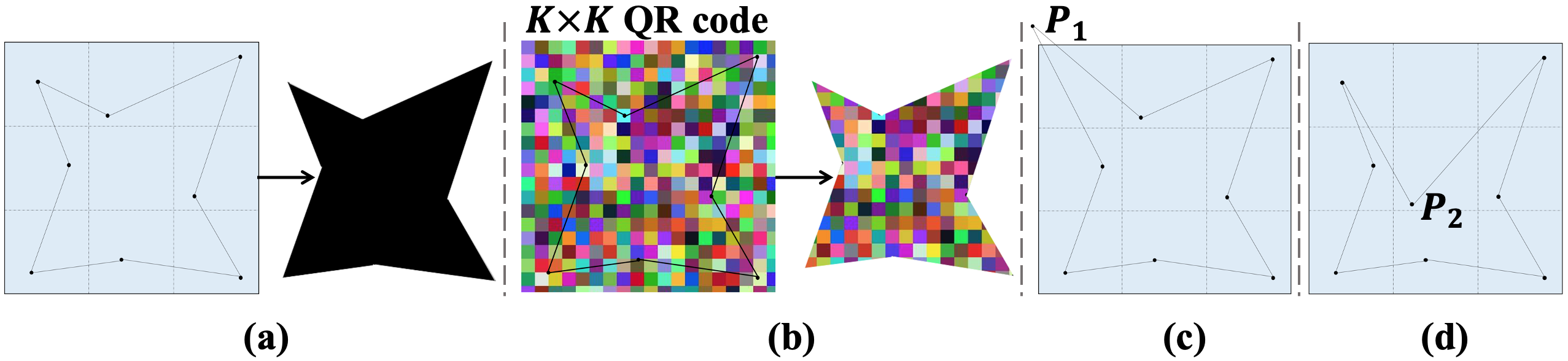} 
\caption{Schematic representation of the unified adversarial patch modeling. }
\vspace{-0.3cm}
\label{figure3}
\end{figure}

\subsection{Generate adversarial sample}

In this study, to effectively maximize the adversarial impact, we define the shape of the adversarial patch as an irregular octagon, with each vertex of the octagon constrained to lie within the outer eight sub-squares of a nine-square lattice. Thus, in the infrared environment, we represent the infrared patch as $P=\{{P}_{1}({i}_{1},{j}_{1}),{P}_{2}({i}_{2},{j}_{2}),...,{P}_{8}({i}_{8},{j}_{8})\}$, and its modeling approach is depicted in Figure \ref{figure3} (a). This modeling method offers the advantage of avoiding vertices of the polygon exceeding the target's boundary (point ${P}_{1}$ in Figure \ref{figure3} (c)) or moving to the center of the target (point ${P}_{2}$ in Figure \ref{figure3} (d)), both of which could weaken the adversarial effect. For the visible patch modeling, as shown in Figure \ref{figure3} (b), we use $K$ to represent the dimension of the color QR code, dividing the bounding box of the target object into $K$ equal parts in the horizontal and vertical directions. In our actual experiments, we set $K$ to 18. $C=\{{C}_{11}({R}_{11},{G}_{11},{B}_{11}),..., {C}_{1K}({R}_{1K},{G}_{1K},{B}_{1K}),...,\\ {C}_{KK}({R}_{KK},{G}_{KK},{B}_{KK})\}$ denotes the color of each square, respectively. Therefore, $P \cap C$ represents the final solved visible light adversarial patch.

Building upon the preceding discussion, we use $\theta(P,P \cap C)$ to denote the cross-modal unified adversarial patch. Therefore, the generation of visible and infrared adversarial samples can be expressed as follows:

\begin{equation}
    \label{Formula2}
    {x}_{vis}^{adv}=S({x}_{vis},\theta(P,P \cap C))
\end{equation}

\begin{equation}
    \label{Formula3}
    {x}_{inf}^{adv}=S({x}_{inf},\theta(P,P \cap C))
\end{equation}
where $S$ represents the linear fusion method, and the adversarial samples are obtained by fusing the clean samples with the cross-modal unified adversarial patch.

In the physical experiment, we crop the printed poster to obtain the irregular polygon color QR code, and the pattern is used to interfere with the visible detector. Cold patches are then deployed on the back of this cropped poster for interference with the infrared detector. To enhance the robustness of the generated unified adversarial patches and to account for potential variations between the digital and physical domains, we introduce Expectation Over Transformation (EOT). EOT has proven to be an effective tool to address domain transfer between different environments, and this approach consists of applying a transform distribution, denoted as $\mathcal{T}$, to model domain transfer between digital and physical domains. The transformation set in $\mathcal{T}$ contains a series of random image transformations, such as view transformation, brightness adjustment, downsampling, etc. By incorporating these different transformations, EOT allows us to model the differences that may arise due to the transition from the digital domain to the physical world. A fundamental advantage of using $\mathcal{T}$ lies in its ability to compensate for small differences in pixel values and positional accuracy that may occur during the simulation of uniform adversarial patches. Therefore, the final expression of the physical domain adversarial sample is as follows:

\begin{equation}
    \label{Formula4}
    {x}_{vis}^{adv}= {\mathbb{E}}_{t \sim \mathcal{T}}t(S({x}_{vis},\theta(P,P \cap C)))
\end{equation}

\begin{equation}
    \label{Formula5}
    {x}_{inf}^{adv}={\mathbb{E}}_{t \sim \mathcal{T}}t(S({x}_{inf},\theta(P,P \cap C)))
\end{equation}

\subsection{Unified adversarial patch attack}

Our objective is twofold: first, to optimize the shape of the infrared patch in the initial stage, obtaining the most adversarial infrared patch shape; and second, in the subsequent stage, to optimize the visible color QR code patch. This optimized patch is then combined with the shape acquired in the first stage, resulting in the most adversarial irregular polygonal color QR code patch. In this work, we consider a realistic attack scenario where the attacker lacks access to internal information about the target model, such as architecture and gradient. Instead, they only possess output information from the model, such as the detection target and its confidence level. Therefore, we treat the confidence of the target as an adversarial loss. The optimization objective is formalized as minimizing the confidence of the detector regarding the target object, expressed as follows:

\begin{equation}
    \label{Formula6}
     \mathop{\arg\min}_{\theta}{\mathbb{E}}_{t \sim \mathcal{T}}{f}_{vis}(t(S({x}_{vis},\theta(P,P \cap C))))
\end{equation}

\begin{equation}
    \label{Formula7}
     \mathop{\arg\min}_{\theta}{\mathbb{E}}_{t \sim \mathcal{T}}{f}_{inf}(t(S({x}_{inf},\theta(P,P \cap C))))
\end{equation}

Building upon the inspiration from HCB \cite{ref34}, we employ PSO \cite{ref37} in this study to independently optimize the adversarial patches in the two distinct stages.

\textbf{Initialization:}
In the optimization process, we commence by randomly generating a population of candidate solutions, consisting of the population ($POP$) and their respective velocity vectors ($V$):

\begin{equation}
    \label{Formula8}
    POP=\{{\theta}_{1},{\theta}_{2},...,{\theta}_{\alpha}\}
\end{equation}

\begin{equation}
    \label{Formula9}
    V=\{{v}_{1},{v}_{2},...,{v}_{\alpha}\}
\end{equation}
where $\alpha$ represents the population size. Each candidate solution in the population POP is denoted as ${\theta}_{a}$, where $a$ ranges from 1 to $\alpha$. Additionally, we use ${v}_{a}$ to represent the direction vector of the particle's motion for ${\theta}_{a}$.

\textbf{Generate adversarial examples.} Each individual, ${\theta}_{a}$, is employed to generate adversarial examples:

\begin{equation}
    \label{Formula10}
    {x}_{a}^{m}=S(x,{\theta}_{a}^{m})
\end{equation}
where $m$ represents the current iteration number of the population. ${x}_{a}^{m}$ denotes the adversarial sample generated by the $a$-th individual in the population of generation $m$.

\textbf{Acquire Individual and Global Optima.} In this phase, we identify both the individual optimal solution and the global optimal solution within the initial population up to the current generation. These solutions will act as directional guides for individual actions in subsequent steps:

\begin{equation}
    \label{Formula11}
    {\theta}_{a,best}^{m}=\mathop{\arg\min}_{{\theta}_{a,best}^{u}}f(x,{\theta}_{a,best}^{u}) \quad \quad u \in [1, m]
\end{equation}

\begin{equation}
    \label{Formula12}
    {\theta}_{best}^{m}=\mathop{\arg\min}_{{\theta}_{a,best}^{u}}f(x,{\theta}_{a,best}^{u}) \quad a \in [1, \alpha] \quad u \in [1, m]
\end{equation}
where ${\theta}_{a,best}^{m}$ and ${\theta}_{best}^{m}$ represent the individual optimal solution and the global optimal solution of $POP$ in generation $m$, respectively.

\textbf{Update Individual Velocity and Position.} To maintain the evolution of the population, we employ the following formula to update the individual speed and position information:

\begin{equation}
    \label{Formula13}
    {v}_{a}^{m+1}=\omega{v}_{a}^{m}+{c}_{1}{r}_{1}({\theta}_{a,best}^{m}-{\theta}_{a}^{m}) + {c}_{2}{r}_{2}({\theta}_{best}^{m}-{\theta}_{a}^{m})
\end{equation}

\begin{equation}
    \label{Formula14}
    {\theta}_{a}^{m+1}={\theta}_{a}^{m}+{v}_{a}^{m+1}
\end{equation}

In the update process, we introduce the following parameters: $\omega$ represents the inertia factor, while ${c}_{1}$ and ${c}_{2}$ denote the learning factors of the particles. Additionally, ${r}_{1}$ and ${r}_{2}$ are random numbers generated from a uniform distribution in the range [0,1].

\begin{algorithm}
	\renewcommand{\algorithmicrequire}{\textbf{Input:}}
	\renewcommand{\algorithmicensure}{\textbf{Output:}}
	\caption{Pseudocode of AdvCL}
	\label{algorithm1}
	\begin{algorithmic}[1]
	
		\REQUIRE Visible image ${x}_{vis}$, Infrared image ${x}_{inf}$, Visible detector ${f}_{vis}$, Infrared detector ${f}_{inf}$, Population size $\alpha$, Iterations $M$, Hyperparameters of PSO: $\omega$,${c}_{1}$,${r}_{1}$,${c}_{2}$, ${r}_{2}$;
		\ENSURE Physical parameters ${\theta}^{\star}$;

        \STATE \textbf{Stage One:}

		\STATE \textbf{Initialization} Randomly set ${POP}_{inf}$, ${V}_{inf}$;

        % \FOR{$m$ $\leftarrow$ 0 to $M$}
        %     \STATE Encoding individual genotype ${\theta}_{g}$;
        % \ENDFOR

        \FOR{$m$ $\leftarrow$ 0 to $M$}
            \FOR{each ${P}_{a,inf}^{m}$ in ${POP}_{inf}^{m}$}
                \STATE ${x}_{a,inf}^{m,adv}={\mathbb{E}}_{t \sim \mathcal{T}}t(S({x}_{inf},{P}_{a,inf}^{m}))$;
                \STATE ${f}_{inf}({x}_{a,inf}^{m,adv}) \rightarrow {P}_{a,inf,best}^{m}, {P}_{inf,best}^{m}$;
            \ENDFOR
            \STATE Update ${v}_{a,inf}^{m+1}$, ${P}_{a,inf}^{m+1}$ by Equation \ref{Formula13}, \ref{Formula14};
        \ENDFOR

        \STATE \textbf{Output:} ${P}_{best}={P}_{inf,best}^{M}$;

        \STATE \textbf{Stage Two:}
        
        \STATE \textbf{Initialization} Randomly set ${POP}_{vis}$, ${V}_{vis}$;

        \FOR{$m$ $\leftarrow$ 0 to $M$}
            \FOR{each ${C}_{a,vis}^{m}$ in ${POP}_{vis}^{m}$}
                \STATE ${x}_{a,vis}^{m,adv}={\mathbb{E}}_{t \sim \mathcal{T}}t(S({x}_{vis},{P}_{best} \cap {C}_{a,vis}^{m}))$;
                \STATE ${f}_{vis}({x}_{a,vis}^{m,adv}) \rightarrow {C}_{a,vis,best}^{m}, {C}_{vis,best}^{m}$;
            \ENDFOR
            \STATE Update ${v}_{a,vis}^{m+1}$, ${C}_{a,vis}^{m+1}$ by Equation \ref{Formula13}, \ref{Formula14};
        \ENDFOR

        \STATE \textbf{Output:} ${\theta}^{\star}=({P}_{best},{P}_{best} \cap {C}_{vis,best}^{M})$;

	\end{algorithmic}  
\end{algorithm}

Algorithm \ref{algorithm1} presents the pseudocode for the proposed TOUAP. The algorithm takes input parameters, including visible light clean sample ${x}_{vis}$, infrared clean sample ${x}_{inf}$, visible detector ${f}_{vis}$, infrared detector ${f}_{inf}$, population size $\alpha$, iteration number $M$, and PSO hyperparameters $\omega$, ${c}_{1}$, ${r}_{1}$, ${c}_{2}$, ${r}_{2}$. These parameters, determined by the attacker, drive the detailed optimization process outlined in Algorithm 1. The final output is the most adversarial unified patch ${\theta}^{\star}$, utilized for subsequent physical attacks.

\section{Experiments}

\subsection{Experimental setting}

\textbf{Dataset.} The FLIR V2\_1 dataset \cite{ref19} serves as the cornerstone for both model training and single-modal digital attack testing in our study. It encompasses a comprehensive collection of 26,442 fully annotated frames, featuring a total of 520,000 bounding box annotations distributed across 15 distinct object categories. This dataset comprises 9,711 thermal images and 9,233 RGB images, captured using a thermal camera and a visible camera mounted on a vehicle. The thermal images were acquired using the Teledyne FLIR Tau 2, boasting a 45-degree horizontal field of view (HFOV) and a 37-degree vertical field of view (VFOV).
For our purposes, we selectively chose infrared and visible images from the FLIR dataset that specifically involve vehicles, ensuring that the width of the vehicle labeling box is greater than or equal to 300 pixels for effective model training. Following a meticulous series of data processing steps, we derived a final dataset named FLIR-TOUAP, consisting of 8,387 infrared images and 3,465 visible images. The FLIR-TOUAP infrared dataset comprises 7,566 samples for training and 821 for testing, while the visible dataset contains 3,116 training samples and 349 testing samples.
The LLVIP dataset \cite{ref20} is a meticulously aligned set of data pairs in both time and space, designed for object detection in the infrared-visible cross modality. This dataset encompasses a total of 30,976 images, organized into 15,488 pairs, featuring 24 night scenes and 2 day scenes. The data acquisition was facilitated by a HIKVISION camera, specifically the HikVision DS-2TD8166BJZFY-75H2F/V2 model. Our utilization of the LLVIP dataset is focused on conducting cross-modal digital attack tests in our research.

\textbf{Object Detector.}
We employ Yolo v3 \cite{ref2} as our chosen object detector for training the infrared vehicle detector. To align with model specifications, we resize the input image to 416×416. To enhance the model's effectiveness during training, we leverage the pre-trained weights of Yolo v3 and fine-tune them on the curated FLIR-TOUAP dataset. Consequently, we achieve an infrared vehicle detector and a visible vehicle detector, boasting average precisions of 92.1\% and 81.6\%, respectively.

\begin{figure}
\centering
\includegraphics[width=1\columnwidth]{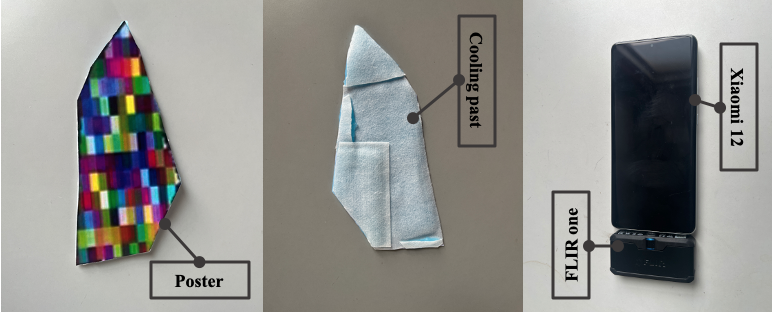} 
\caption{Experimental devices. Printed posters serve as physical patches in the visible light environment, while cold patches are employed as physical patches in the infrared environment. The imaging equipment utilized includes a Xiaomi 12 mobile phone and a FLIR One thermal imager.}
\vspace{-0.3cm}
\label{figure4}
\end{figure}

\textbf{Experimental devices.}
The experimental devices employed in this study is illustrated in Figure \ref{figure4}, comprising a unified adversarial patch, a Xiaomi 12 phone, and an infrared thermal imager. The front of the unified adversarial patch features a printed and cropped poster, while the back is equipped with cold patches, capable of sustained cooling to 24 degrees Celsius for up to 4 hours. Regarding the camera equipment, it has been verified that the efficacy of the proposed TOUAP remains independent of the specific camera model utilized. Key parameters of the FLIR One thermal imager include an infrared resolution of 160×120 and a visible resolution of 1440×1080, with an infrared NETD\textless70mk.

\textbf{Evaluation Metrics.}
The objective of our proposed TOUAP is to deceive both infrared and visible vehicle detectors, causing them to either miss or misidentify the target object. We employ the Attack Success Rate (ASR) as a metric to assess the adversarial performance of TOUAP. A higher ASR indicates a more effective adversarial impact of TOUAP, and vice versa. The formula is expressed as follows:

\begin{equation}
\label{eq:Positional Encoding}
\begin{split}
    &{\rm ASR}(X) = 1-\frac{1}{N}\sum_{i=1}^{N}F({f}_{vis}({x}_{vis}^{adv}),{f}_{inf}({x}_{inf}^{adv}))\\
    & \begin{split}
    &F({f}_{vis}({x}_{vis}^{adv}),{f}_{inf}({x}_{inf}^{adv}))=\\
       & \begin{cases}
        1 & {f}_{vis}({x}_{vis}^{adv}) < \delta \quad and \quad {f}_{inf}({x}_{inf}^{adv}) < \delta \\
        0 & otherwise
        \end{cases}
    \end{split}
\end{split}
\end{equation}
where $N$ represents the number of true positive labels in the dataset $X$ that can be detected by the detector in the absence of an attack. In all our attack experiments, the threshold is set to $\delta$=0.5.

Competing methods. The proposed TOUAP is categorized as a black-box attack. For a fair comparison, we select other black-box attacks, namely UnfAP \cite{ref16} and HCB \cite{ref34}, as baselines for experimental comparison.

Other details. We set the population size to $\alpha$=100 and the number of iterations to $M$=10. The hyperparameters of the PSO are set as follows: $\omega$=0.9, ${c}_{1}$=1.6, ${r}_{1}$=0.5, ${c}_{2}$=1.4, ${r}_{2}$=0.5. All attack experiments are conducted on an NVIDIA GeForce RTX 4090 GPU.

\subsection{Evaluation of effectiveness}

\textbf{Digital attacks.} To comprehensively assess the effectiveness of the proposed TOUAP in the digital environment, we conduct single-modal digital attack tests on FLIR-TOUAP. For the visible modal attack, we concurrently optimize the shape of the patch and the color QR code. After confirming the effectiveness of our method in single-modal attacks, we deploy TOUAP to conduct cross-modal attacks on the LLVIP dataset. The results of our experiments are presented in Table \ref{Table1}, showcasing superior adversarial effects in single-modal digital attacks for both the infrared and visible modalities, achieving a nearly 100\% attack success rate with a low query cost. In the cross-modal attack, we achieve an attack success rate of 98.17\%, with an average query number of 43.38. This underscores the effectiveness of TOUAP in the digital environment and establishes a foundation for subsequent physical experiments. Table \ref{Table2} displays the experimental results of our method and the baselines, demonstrating the greater adversarial efficacy of our approach compared to baseline methods.

\begin{table}
\centering
\setlength{\belowcaptionskip}{0.3cm}
\caption{\label{Table1} Deploying TOUAP on FLIR-TOUAP and LLVIP to perform digital attacks.}
\begin{tabular}{cccccc}
\hline
\multicolumn{2}{c}{Infrared-modal} & \multicolumn{2}{c}{Infrared-modal} & \multicolumn{2}{c}{Infrared-modal}\\

\hline
ASR & Query & ASR & Query & ASR & Query \\

\hline

99.68&6.08&97.73&74.22&98.17&43.38\\

\hline
\end{tabular}
\end{table}

\begin{table}
\centering
\setlength{\belowcaptionskip}{0.3cm}
\caption{\label{Table2} Comparison of experimental results between the proposed method and the baseline methods.}
\begin{tabular}{cccc}
\hline
Method & Ours & UnfAP & HCB\\

\hline
ASR &98.17&89.73&21.79 \\

\hline

Query &43.38&198.47&907.71\\

\hline
\end{tabular}
\end{table}

\begin{figure}
\centering
\includegraphics[width=1\columnwidth]{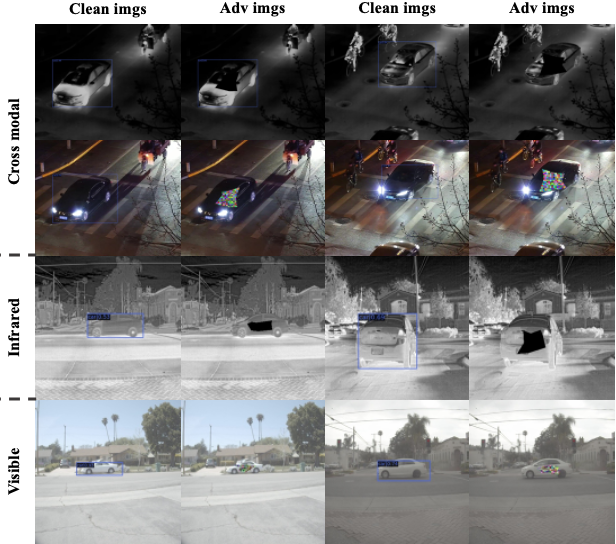} 
\caption{Single-modal/cross-modal digital adversarial examples generated by TOUAP.}
\vspace{-0.3cm}
\label{figure5}
\end{figure}

We present the single-modal and cross-modal digital samples generated by TOUAP in Figure \ref{figure5}. In the case of single-modal adversarial samples, it is evident that although the patch is easily detectable, it does not compromise the semantic information of the vehicle, yet it effectively deceives the target detector. As for the cross-modal adversarial samples, it is noticeable that the shape and position of the visible patch and the infrared patch align (only patches of different modals with the same position and shape can be deployed as a unified adversarial patch in the physical environment). This consistency allows for the simultaneous successful attack on both the infrared and visible vehicle detectors.

\begin{table}
\centering
\setlength{\belowcaptionskip}{0.3cm}
\caption{\label{Table3} Experimental results of physical attacks (\%).}
\begin{tabular}{ccccccc}
\hline
~ & ${-30}^{\circ}$ & ${-15}^{\circ}$ & ${0}^{\circ}$ &${15}^{\circ}$&${30}^{\circ}$&Overall\\

\hline
Infrared&100&100&100&100&100&100 \\

\hline

Visible &50&100&100&100&81.80&87.50\\

\hline
\end{tabular}
\end{table}

\begin{figure}
\centering
\includegraphics[width=1\columnwidth]{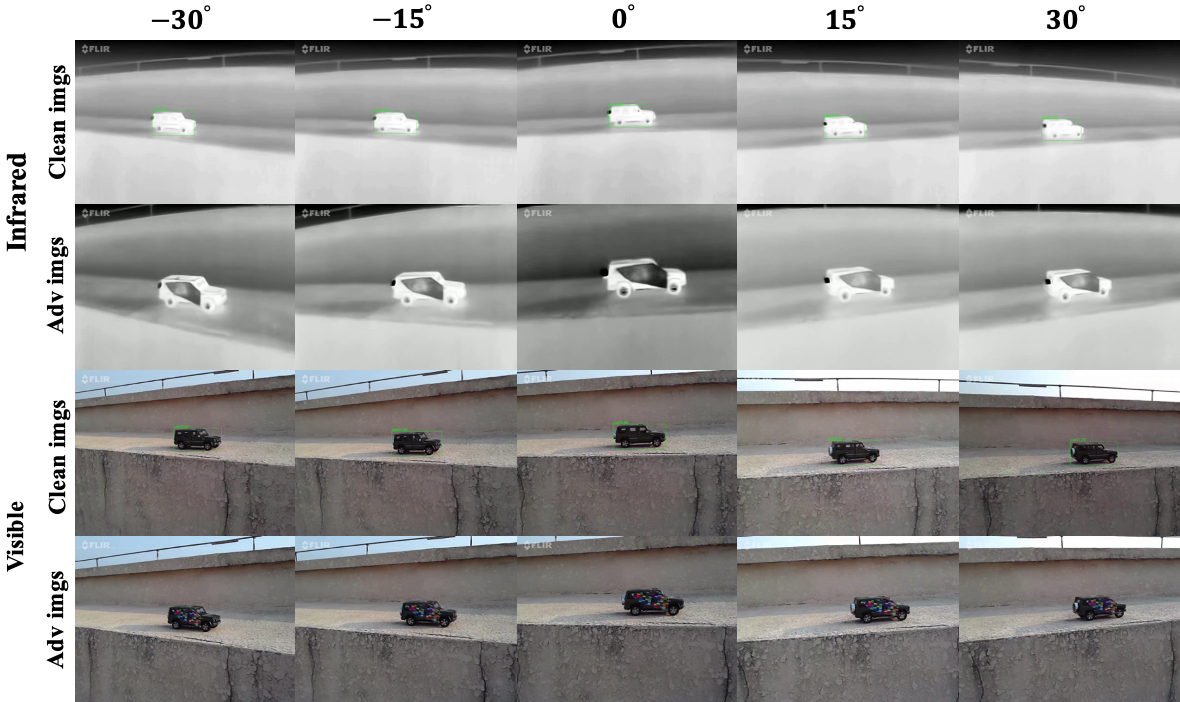} 
\caption{Physical samples generated by TOUAP. }
\vspace{-0.3cm}
\label{figure6}
\end{figure}

Physical attacks. Building upon the insights gained from AIP experiments and discussions, where frontal deployment showed some degree of antagonism within the ${30}^{\circ}$ perspective, our physical experiments with TOUAP assess its adversarial impact under five viewpoints: ${-30}^{\circ}$, ${-15}^{\circ}$, ${0}^{\circ}$, ${15}^{\circ}$, and ${30}^{\circ}$. The summarized experimental results in Table \ref{Table3} lead to two key conclusions: (1) Overall, our unified adversarial patches achieve 100\% and 87.50\% attack success rates against the infrared detector and visible light detector, respectively, validating the effectiveness of the proposed method in executing cross-modal attacks; (2) The unified adversarial patch generated by TOUAP attains a 100\% attack success rate against the infrared pedestrian detector from all viewing angles. Furthermore, the unified adversarial patch secures a 100\% attack success rate against the visible light detector within a small viewing angle (within the range of plus or minus ${15}^{\circ}$). Refer to the Supplementary material for our video presentation. Figure \ref{figure6} showcases physical samples generated by the proposed method, highlighting its efficacy in executing effective physical attacks against infrared and visible light detectors at various viewing angles.

In summary, the experimental results presented in Tables \ref{Table3} and \ref{Table3} demonstrate that our method is capable of achieving superior cross-modal attacks in both digital and physical environments, thereby validating the effectiveness of TOUAP.

\subsection{Evaluation of robustness}

In this study, we conduct a comprehensive series of experiments to assess the effectiveness of the proposed TOUAP in attacking advanced object detectors within a challenging black-box setting. Our targets included five widely used advanced detectors: DETR \cite{ref38}, Mask Rcnn \cite{ref39}, Faster Rcnn \cite{ref40}, Libra Rcnn \cite{ref41}, and RetinaNet \cite{ref42}. To ensure a fair evaluation, we fine-tune these pre-trained models on the FLIR-TOUAP dataset, achieving impressive mean average precision (mAP) scores on the test set for infrared detectors, with values of 94.6\%, 94.2\%, 94.4\%, 95.6\%, and 95.5\%, respectively. In the visible light environment, the mAP scores are 87.0\%, 73.7\%, 73.4\%, 87.4\%, and 75.5\%, respectively. The experimental results are summarized in Table \ref{Table4}, leading to the following key conclusions: (1) The proposed method, on average, achieves an ASR of 93.53\%, 90.20\%, and 86.99\% against advanced infrared, visible, and cross-modal detectors, respectively, demonstrating the effectiveness of TOUAP in attacking advanced detectors and confirming its robustness; (2) Our method effectively paralyzes all models, except DETR, in the infrared modal attack, achieving a 100\% attack success rate for Mask Rcnn and Faster Rcnn; (3) TOUAP exhibits efficient adversarial effects, with an average query number of less than 200; (4) DETR exhibits greater robustness compared to other models, as indicated by its lower attack success rate when targeted by TOUAP. This observation suggests that transformer-based models may possess enhanced resilience.

\begin{table}
\centering
\setlength{\belowcaptionskip}{0.3cm}
\caption{\label{Table4} Evaluation across various detectors.}
\begin{tabular}{ccccccc}
\hline
\multicolumn{1}{c}{\multirow{2}{*}{Model}}&\multicolumn{2}{c}{Infrared-modal} & \multicolumn{2}{c}{Infrared-modal} & \multicolumn{2}{c}{Infrared-modal}\\

\cmidrule(r){2-3}%短横线
\cmidrule(r){4-5}%短横线
\cmidrule(r){6-7}%短横线
\multicolumn{1}{c}{}&ASR & Query & ASR & Query & ASR & Query \\

\hline

Yolo&99.68&6.08&97.73&74.22&98.17&43.38\\

\hline

DETR&62.35&478.86&61.11&403.89&50.00&579.63\\

\hline

Mask&100&12.41&100&12.79&98.94&14.08\\

\hline

Faster&100&2.49&100&15.2&85.29&150.32\\

\hline

Libra&99.48&19.21&90.91&95.43&92.12&87.68\\

\hline

Retina&99.64&31.76&91.43&107.83&97.41&63.82\\

\hline

AVG&93.53&91.80&90.20&118.23&86.99&156.49\\

\hline
\end{tabular}
\end{table}

In summary, the experimental findings presented in Table \ref{Table2} underscore the superior adversarial nature of our method compared to the baselines. Furthermore, the results in Table \ref{Table4} affirm the efficacy and robustness of the proposed TOUAP against diverse advanced detectors. These outcomes serve as compelling evidence for the resilience of TOUAP. Given its superior performance over the baselines, we advocate for increased attention to the proposed method.

% \section{Discussion}

\subsection{Ablation study}

In this section, we delve into the physical parameters of TOUAP, specifically examining the impact of the irregular polygon's shape and the dimension $K$ of the colored QR code.

\textbf{Ablation Experiments on Shapes.}
For our uniform adversarial patch, we employ an irregular polygon, with the positions of vertices constrained to the outer eight subsquares of the nine-square grid. We conduct ablation experiments, varying the shape from a triangle to an octagon, to assess their influence on cross-modal digital attacks. Results in Table \ref{Table5} reveal that the attack success rate generally rises with the number of polygon vertices. Even with a simple triangle, our method achieves an ASR of up to 87.68\%. The ASR growth saturates when the vertex count reaches 4, indicating a slower increase afterward.

\textbf{Ablation Learning of $K$.}
In this analysis, we explore the impact of the dimension $K$ on the cross-modal attack, varying its value from 1 to 30. Results in Table \ref{Table6} demonstrate a general increase in ASR with higher $K$ values. Remarkably, even with $K$=1, representing a monochromatic unified adversarial patch, an ASR of up to 90.00\% is attained. The ASR stabilizes at $K$=6, with a gradual increase as $K$ further grows.

\begin{table}
\centering
\setlength{\belowcaptionskip}{0.3cm}
\caption{\label{Table5} Ablation study of shape.}
\begin{tabular}{ccccccc}
\hline
$Shape$&3&4&5&6&7&8\\

\hline
ASR (\%)&87.68&97.38&96.99&97.64&98.17&98.17\\

\hline

Query&159.27&51.93&57.91&47.74&37.56&43.38\\

\hline
\end{tabular}
\end{table}

\begin{table}
\centering
\setlength{\belowcaptionskip}{0.3cm}
\caption{\label{Table6} Ablation study of K.}
\begin{tabular}{ccccccc}
\hline

$K$&1&6&12&18&24&30\\

\hline
ASR (\%)&90.00&97.83&98.26&98.17&98.26&96.52\\

\hline

Query&106.21&34.71&28.28&43.38&30.83&41.03\\

\hline
\end{tabular}
\end{table}

\subsection{Defense of TOUAP}

In this section, we explore defense strategies against TOUAP, focusing on adversarial training \cite{AT} and DW \cite{DW}.
\textbf{Adversarial training.}
To enhance robustness, we perform adversarial training by fine-tuning the Yolo v3 model using data augmentation, maintaining a 5:1 ratio of adversarial to clean samples in the training dataset. Following adversarial training, we obtain infrared/visible Yolo v3 models with accuracies of 89.3\% and 93.8\%, respectively.
\textbf{DW.}
In the non-blind setting of DW, where the defender possesses information about the disturbance's position, we apply image inpainting separately to the visible light and infrared adversarial samples. The defense is deemed successful if either of the infrared or visible light detectors correctly identifies the adversarial samples. Results in Table 7 show that:
(1) Adversarial training is effective, enhancing detector robustness against TOUAP.
(2) Non-blind DW proves to be a potent defense against TOUAP, outperforming adversarial training.
(3) Despite their effectiveness, neither adversarial training nor DW achieves complete defense against TOUAP.

\begin{table}
\centering
\setlength{\belowcaptionskip}{0.3cm}
\caption{\label{Table7} Adversarial defenses against TOUAP.}
\begin{tabular}{ccccccc}
\hline
\multicolumn{1}{c}{\multirow{2}{*}{Method}}&\multicolumn{2}{c}{Infrared-modal} & \multicolumn{2}{c}{Infrared-modal} & \multicolumn{2}{c}{Infrared-modal}\\

\cmidrule(r){2-3}%短横线
\cmidrule(r){4-5}%短横线
\cmidrule(r){6-7}%短横线
\multicolumn{1}{c}{}&ASR & Query & ASR & Query & ASR & Query \\

\hline

No defense&99.68&6.08&97.73&74.22&98.17&43.38\\

\hline

Adv training&73.63&313.37&76.63&309.22&68.31&401.30\\

\hline

DW&57.69&-&44.87&-&33.72&-\\

\hline

\end{tabular}
\end{table}

\section{Conclusion}

In this work, we introduce TOUAP, a two-stage optimized unified adversarial patch designed for physical attacks on visible-infrared cross-modal vehicle detectors. The first stage employs a nine-palace grid to model the infrared adversarial patch, optimizing it with PSO. Subsequently, in the second stage, the patch is modeled based on a color QR code and further optimized using PSO. The resulting patch shape from the first stage is combined with the color QR code, yielding the irregular polygon visible adversarial patch.
Extensive experiments demonstrate the efficacy and robustness of TOUAP. In terms of effectiveness, the method achieves no less than an 80\% ASR against the cross-modal Yolo v3 detector in both digital and physical environments, affirming its effectiveness. Regarding robustness, TOUAP not only surpasses baseline attack effects but also efficiently executes cross-modal attacks against various advanced detectors, confirming its robustness. Given the superior performance of TOUAP, we advocate for broader attention to this approach.
Future research will continue to explore cross-modal attacks, such as physical attacks against visible-radar detectors and visible-infrared-radar detectors. Additionally, investigating more effective defense methods against cross-modal physical attacks will be a key focus.

% \section*{Acknowledgment}

% The preferred spelling of the word ``acknowledgment'' in America is without 
% an ``e'' after the ``g''. Avoid the stilted expression ``one of us (R. B. 
% G.) thanks $\ldots$''. Instead, try ``R. B. G. thanks$\ldots$''. Put sponsor 
% acknowledgments in the unnumbered footnote on the first page.

% \section*{References}

% Please number citations consecutively within brackets \cite{b1}. The 
% sentence punctuation follows the bracket \cite{b2}. Refer simply to the reference 
% number, as in \cite{b3}---do not use ``Ref. \cite{b3}'' or ``reference \cite{b3}'' except at 
% the beginning of a sentence: ``Reference \cite{b3} was the first $\ldots$''

% Number footnotes separately in superscripts. Place the actual footnote at 
% the bottom of the column in which it was cited. Do not put footnotes in the 
% abstract or reference list. Use letters for table footnotes.

% Unless there are six authors or more give all authors' names; do not use 
% ``et al.''. Papers that have not been published, even if they have been 
% submitted for publication, should be cited as ``unpublished'' \cite{b4}. Papers 
% that have been accepted for publication should be cited as ``in press'' \cite{b5}. 
% Capitalize only the first word in a paper title, except for proper nouns and 
% element symbols.

% For papers published in translation journals, please give the English 
% citation first, followed by the original foreign-language citation \cite{b6}.

\bibliographystyle{IEEEtran}
\bibliography{IEEEfull}

% \begin{thebibliography}{00}
% \bibitem{b1} G. Eason, B. Noble, and I. N. Sneddon, ``On certain integrals of Lipschitz-Hankel type involving products of Bessel functions,'' Phil. Trans. Roy. Soc. London, vol. A247, pp. 529--551, April 1955.
% \bibitem{b2} J. Clerk Maxwell, A Treatise on Electricity and Magnetism, 3rd ed., vol. 2. Oxford: Clarendon, 1892, pp.68--73.
% \bibitem{b3} I. S. Jacobs and C. P. Bean, ``Fine particles, thin films and exchange anisotropy,'' in Magnetism, vol. III, G. T. Rado and H. Suhl, Eds. New York: Academic, 1963, pp. 271--350.
% \bibitem{b4} K. Elissa, ``Title of paper if known,'' unpublished.
% \bibitem{b5} R. Nicole, ``Title of paper with only first word capitalized,'' J. Name Stand. Abbrev., in press.
% \bibitem{b6} Y. Yorozu, M. Hirano, K. Oka, and Y. Tagawa, ``Electron spectroscopy studies on magneto-optical media and plastic substrate interface,'' IEEE Transl. J. Magn. Japan, vol. 2, pp. 740--741, August 1987 [Digests 9th Annual Conf. Magnetics Japan, p. 301, 1982].
% \bibitem{b7} M. Young, The Technical Writer's Handbook. Mill Valley, CA: University Science, 1989.
% \end{thebibliography}
% \vspace{12pt}
% \color{red}
% IEEE conference templates contain guidance text for composing and formatting conference papers. Please ensure that all template text is removed from your conference paper prior to submission to the conference. Failure to remove the template text from your paper may result in your paper not being published.

\end{document}